\documentclass[a4paper,conference]{IEEEtran}
\usepackage{cite}

\ifCLASSINFOpdf
   \usepackage[pdftex]{graphicx}
   \graphicspath{{../png/}{../jpeg/}}
   \DeclareGraphicsExtensions{.pdf,.jpeg,.png}
\else
   \usepackage[dvips]{graphicx}
   \graphicspath{{../eps/}}
   \DeclareGraphicsExtensions{.eps}
\fi
\usepackage{amsmath}
\usepackage{url}

\usepackage{indentfirst}
\usepackage{multirow}
\usepackage{color}
\usepackage{hyperref}
\hypersetup{hidelinks} 

\begin{document}
%
\title{A lightweight multi-scale context network for salient object detection in optical remote sensing images}

%

\author{\IEEEauthorblockN{Yuhan Lin\IEEEauthorrefmark{1},
Han Sun\IEEEauthorrefmark{1},
Ningzhong Liu\IEEEauthorrefmark{1},
Yetong Bian\IEEEauthorrefmark{1}, 
Jun Cen\IEEEauthorrefmark{1}, and 
Huiyu Zhou\IEEEauthorrefmark{2}
} 
\IEEEauthorblockA{\IEEEauthorrefmark{1} College of Computer Science and Technology, Nanjing University of Aeronautics and Astronautics 
Nanjing, China \\
MIIT Key Laboratory of Pattern Analysis and Machine Intelligence, Nanjing, China \\
Email: sunhan@nuaa.edu.cn}
\IEEEauthorblockA{\IEEEauthorrefmark{2}School of Computing and Mathematical Sciences, University of Leicester, Leicester LE1 7RH, UK}}


\maketitle

\begin{abstract}
Due to the more dramatic multi-scale variations and more complicated foregrounds and backgrounds in optical remote sensing images (RSIs), the salient object detection (SOD) for optical RSIs becomes a huge challenge. However, different from natural scene images (NSIs), the discussion on the optical RSI SOD task still remains scarce. In this paper, we propose a multi-scale context network, namely MSCNet, for SOD in optical RSIs. Specifically, a multi-scale context extraction module is adopted to address the scale variation of salient objects by effectively learning multi-scale contextual information. Meanwhile, in order to accurately detect complete salient objects in complex backgrounds, we design an attention-based pyramid feature aggregation mechanism for gradually aggregating and refining the salient regions from the multi-scale context extraction module. Extensive experiments on two benchmarks demonstrate that MSCNet achieves competitive performance with only 3.26M parameters. The code will be available at https://github.com/NuaaYH/MSCNet.
\end{abstract}



%
\IEEEpeerreviewmaketitle

\section{Introduction}
Salient object detection (SOD) aims at modeling the visual attention mechanism of human beings \cite{borji2019salient}\cite{wang2021salient}, which can accurately locate and segment the most attractive object in the image, and is widely used as a pre-processing procedure for down-streaming visual tasks, such as image segmentation \cite{fan2018associating}, object recognition \cite{ren2013region}, object tracking \cite{zhou2021saliency}, image retrieval \cite{mohedano2018saliency} and so on.

During the past decades, hundreds of traditional SOD methods rely on the hand-crafted features, which are far inferior to the current Full Convolutional Network (FCN) based methods. Although remarkable achievements of FCN based SOD models for natural scene images (NSIs) have been made in recent decades, there is only a limited amount 
of researches focusing on SOD for optical remote sensing images (RSIs). Optical RSIs refer to the images captured by aircrafts or satellite equipments from a high altitude. Specifically, there are three main challenging problems in the optical RSI SOD task. First, due to the larger scale variations, the RSI SOD models need to be very sensitive to the scale changes of the salient object. Second, slender salient objects such as rivers tend to span the entire image frame, which brings more difficulties to the complete detection of such objects. Third, there are more challenges in complex backgrounds. Because RSIs are collected from a bird’s-eye view of the ground with flexible distance, RSIs contain more complicated backgrounds such as shadows and confusing surroundings.

Due to above issues, the current methods designed for NSI SOD are not well suited to RSI SOD task. In the past two years, some meaningful methods are specially designed to solve the RSI SOD task. LVNet \cite{li2019nested} outperforms many state-of-the-art SOD methods in optical RSI datasets through a two-stream pyramid multi-scale module and an encoder–decoder module with nested connections. DAFNet \cite{zhang2020dense} is 
inspired by the powerful effects of the attention mechanism and designs the dense attention fluid structure to combine multi-level attention cues, which achieves excellent performance. Although these methods have made significant contributions to the RSI SOD task, there still remain some issues, i.e., heavy model parameters and incomplete salient objects.               

In this work, we aim at designing a lightweight multi-scale context network based on MobileNet V2 \cite{sandler2018mobilenetv2} for accurate salient object detection in optical RSIs, called MSCNet. For better feature extraction of salient objects, we propose a multi-scale context extraction (MSCE) module to solve the scale variation problem, and build a multi-scale feature decoder based on this module. Different from those U-Net \cite{ronneberger2015u} based model that generates the saliency map from the last layer of the symmetric decoder, we adopt an attention-based pyramid feature aggregation (APFA) mechanism to gradually fuse multi-layer feature maps in a form similar to an inverted pyramid. At last, we evaluate the effectiveness of the proposed lightweight network.

\section{Related Work}
\subsection{Salient Object Detection for NSIs}
At early stages, many methods \cite{zhao2015saliency, wang2015deep, li2015visual} are mainly based on fully connected layers to distinguish local or global saliency. Although the accuracy of these methods far exceeds that of traditional models based on hand-crafted features, there are still some problems that cannot well capture key spatial information and shortcomings of time overhead. Later, the Full Convolutional Network (FCN) \cite{long2015fully} architecture, which achieves excellent results in semantic segmentation, becomes the mainstream framework of salient object detection. Pang et al. \cite{pang2020multi} proposes the aggregate interaction module to fuse the adjacent-scale features of the encoder and the self-interaction module to further extract multi-scale context information. Zhao et al. \cite{zhao2019pyramid} presents a spatial attention module for low-level features to suppress background noise and highlight salient areas, and a channel attention module for high-level features to select effective semantic information. And Sun et al. \cite{sun2021multi} uses the first and the last feature map of the encoder to learn the edge information of the salient object, and then the edge features are incorporated into the decoding process as supplementary information. 

With the continuous improvement of the SOD network, the blind pursuit of accuracy makes the model size larger and larger. And some models can reach tens of millions of parameters in size, e.g., 176.3M for U2Net \cite{qin2020u2}, so the request of lightweight model arises. Liu et al. \cite{liu2021samnet} uses the stereo attention multi-scale (SAM) module to achieve comparable performance with the state-of-the-art methods, which only contains 1.33M parameters. Gao et al. \cite{gao2020highly} designs a novel dynamic weight decay strategy to learn the number of 
channels in each scale of gOctConvs, thereby reducing nearly 80\% of the parameters, while the performance degradation is negligible. Huang et al. \cite{huang2020lightweight} proposes a lightweight bottleneck block to improve accuracy and efficiency, which consists of linear bottlenecks, inverted residuals and separable convolution in the depth direction.

\subsection{Salient Object Detection for RSIs}
Thanks to the first RSI SOD dataset named ORSSD constructed, related deep learning models for RSI SOD task have emerged and developed in the last two years. Li et al. \cite{li2019nested} proposes an L-shaped module to solve the problem of scale diversity and a V-shaped module to suppress cluttered backgrounds and highlight salient objects. Zhang et al. \cite{zhang2020dense} designs an end-to-end dense attention flow network, including the dense attention flow (DAF) structure and the global context-aware attention (GCA) mechanism. Cong et al. \cite{cong2021rrnet} presents a relational reasoning network with parallel multi-scale attention, including a relational reasoning module and a parallel multi-scale attention module. The above models rely on heavy network modules to improve the performance of SOD, but ignore the requirement for embedding RSI SOD models into lightweight mobile devices in the most real-world applications. 

\section{Approach}
In this section, we first give an overview of the proposed multi-scale context network (MSCNet), which is shown in Fig. 1. Then, the multi-scale context extraction module (MSCE) and  the attention-based pyramid feature aggregation mechanism (APFA) are discussed in detail.  Finally, loss functions are explained clearly.

\subsection{Overview}
As shown in Fig. 1, the overall framework mainly contains the encoder, multi-scale context extraction (MSCE) module and attention-based pyramid feature aggregation (APFA) mechanism. The encoder adopts MobileNet V2 as its backbone network. Specifically, the MobileNet V2 network can be grouped into five blocks by a series of downsampling operations with a stride of 2. The outputs of these blocks are used as the multi-level feature maps from Conv-1 to Conv-5. Then the feature maps are fed into MSCE module to explore more effective multi-scale contextual information, and then four sets of feature maps that indicate the regions of salient objects are obtained. At last, these four groups of feature maps are processed by APFA to generate the final predicted saliency map.
\begin{figure*}[htbp]
  \centering
  \includegraphics[width=4.8in]{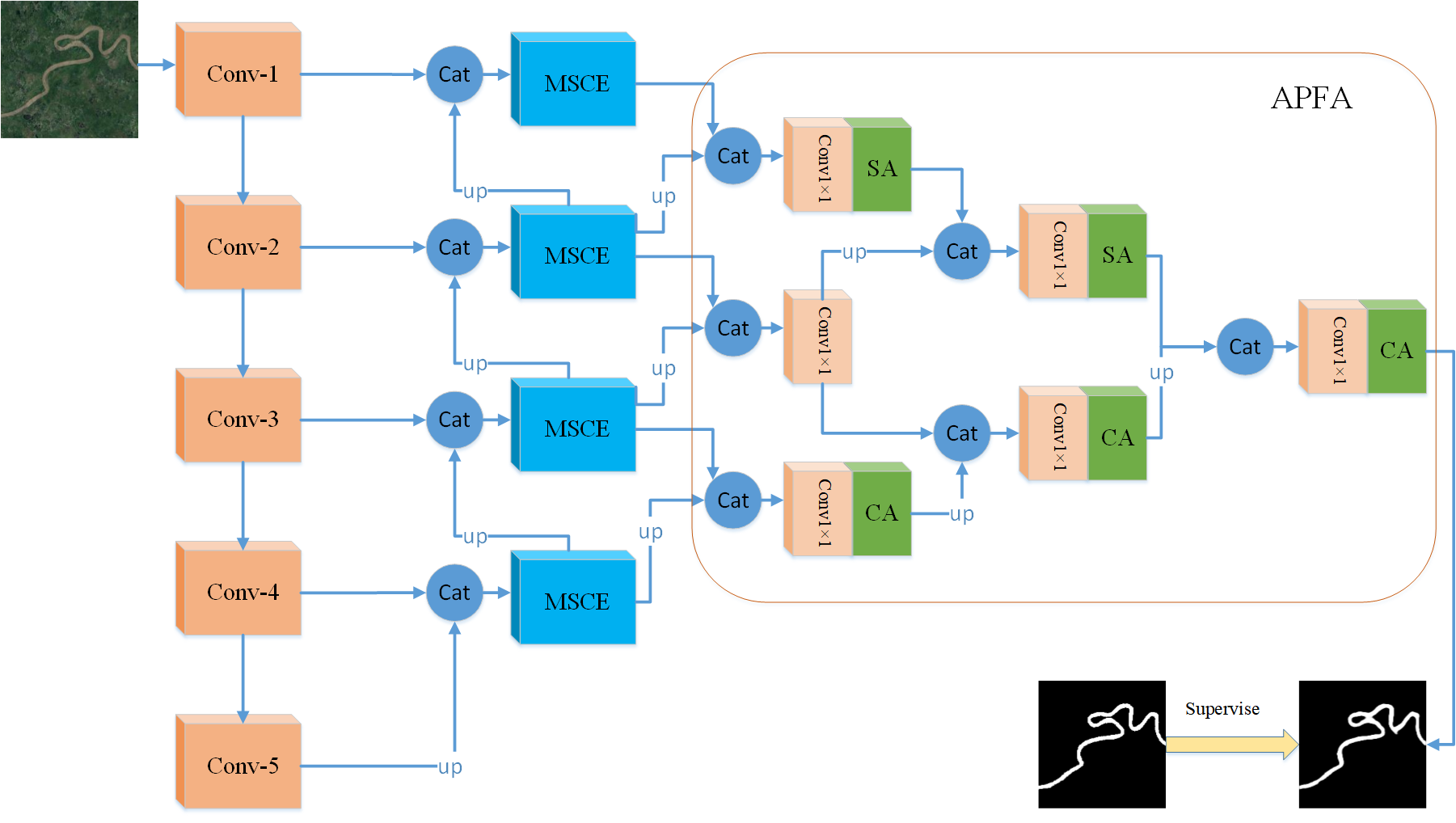}
  \caption{ The architecture of MSCNet. CA means the channel attention module, SA means the spatial attention module and up stands for upsampling operation with scale factor of 2.} 
\end{figure*}
\begin{figure}[h]
  \centering
  \includegraphics[width=2.8in]{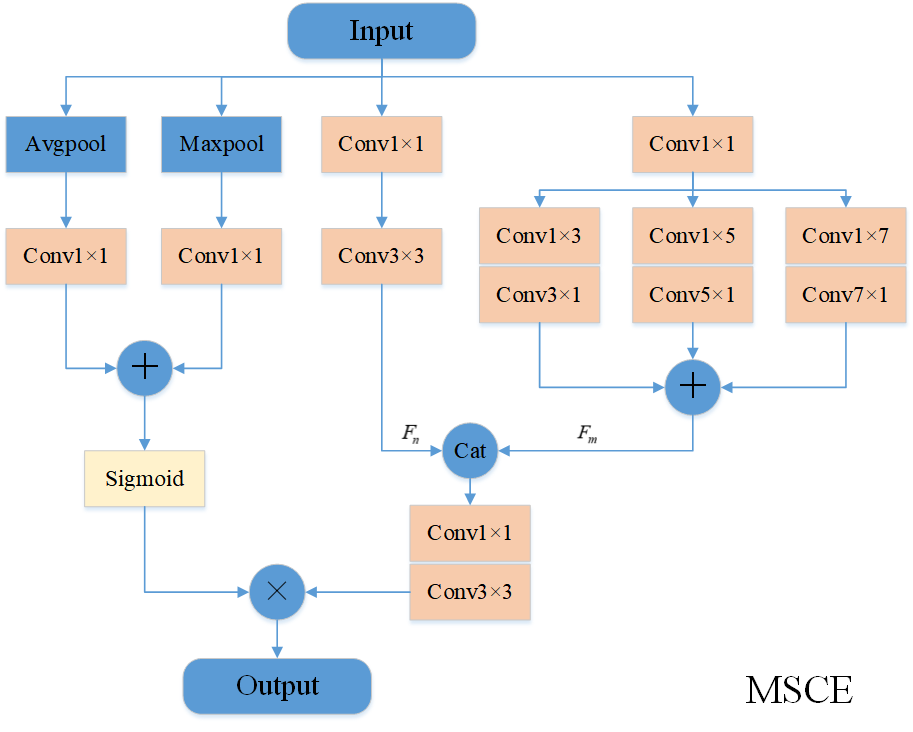}
  \caption{The architecture of MSCE.} 
\end{figure}

\subsection{Multi-scale context extraction (MSCE) module}
Many existing multi-scale feature extraction modules have been proved to enhance the capability of multi-scale object detection to some extent. Motivated by these classic methods such as PPM \cite{zhao2017pyramid}, ASPP \cite{chen2017deeplab} and Inception \cite{szegedy2016rethinking}, we propose a multi-scale context extraction (MSCE) module to solve the severe problem of  multi-scale variation in optical RSIs. And then the multi-scale feature decoder based on the MSCE module is adopted to generate four groups of feature maps.

Firstly, we discuss the details of the MSCE module, shown in  Fig. 2. Specifically, there are two key components in this module. The input feature maps are processed by two different $1 \times 1$ convolutional layers to obtain two feature maps. One feature map is refined 
with conventional $3 \times 3$ convolution, while the other feature map is fed into three independent parallel asymmetric convolution branches for multi-scale feature extraction. The size of the convolution kernels used in the three asymmetric convolution branches are 3, 5, and 7, respectively. 
This process can be formulated as:
\begin{equation}
  f_1  = C{\rm{onv}}_{{\rm{3 \times 3}}} (C{\rm{onv}}_{1{\rm{ \times }}1} (f))
\end{equation}
\begin{equation}
  \begin{split}
  f_2  = AConv_{3{\rm{ \times }}3} (\widetilde{f}) + AConv_{{\rm{5 \times 5}}} (\widetilde{f}) + AConv_{{\rm{7 \times 7}}} (\widetilde{f})\\
  ,\widetilde{f} = Conv_{{\rm{1 \times 1}}} (f)
  \end{split}
\end{equation}
 where $f$ denotes the input feature map, $Conv_{1 \times 1}$ denotes the $1 \times 1$ convolutional layer,  $Conv_{3 \times 3}$ denotes the $3 \times 3$ convolutional layer, $AConv_{i \times i}$ is the asymmetric convolution  in which $i$ denotes the size of convolution kernel. All the convolutional layer are  followed by the batch normalization and ReLU activation.

 In addition, we design a channel reweighting branch. In this branch, the input feature map is compressed into a one-dimensional vector by the maximum pooling and average pooling  operations. Then the two one-dimensional vectors are fed into a fully-connected layer 
 implemented by $1 \times 1$ convolution and further combined to generate a channel attention vector. This process can be formulated as:
 \begin{equation}
   Atten = \sigma (fc(avgpool(f)) + fc(maxpool(f)))
\end{equation}
where $\sigma $ denotes the sigmoid function, $fc$ denotes the fully-connected layer, $avgpool$ and $maxpool$ are the average-pooling and max-pooling in spatial dimension, respectively.

Next, the structure of the multi-scale feature decoder is described in detail. Based on our proposed MSCE module, we construct a multi-scale feature decoder consisting of four MSCE modules. For example, the deepest feature map of the encoder is upsampled twice and combined with the corresponding encoder feature map of the same resolution size in the upper layer. The first multi-scale contextual feature map is obtained by feeding above combined feature map to the first MSCE module and then used as a new deep layer feature to continue fusing with other encoder feature maps until four groups of feature maps of different resolution sizes are generated. The process in each MSCE module can be formulated as:
\begin{equation}
  F = Atten \odot Conv_{3 \times 3} (Conv_{1 \times 1} (Cat(f_1 ,f_2)))
\end{equation}
where $ \odot $ denotes element-wise multiplication, $Cat$ denotes feature concatenation along channel 
axis and $F$ denotes the output feature map of MSCE.
\begin{figure}[h]
  \centering
  \includegraphics[width=3in]{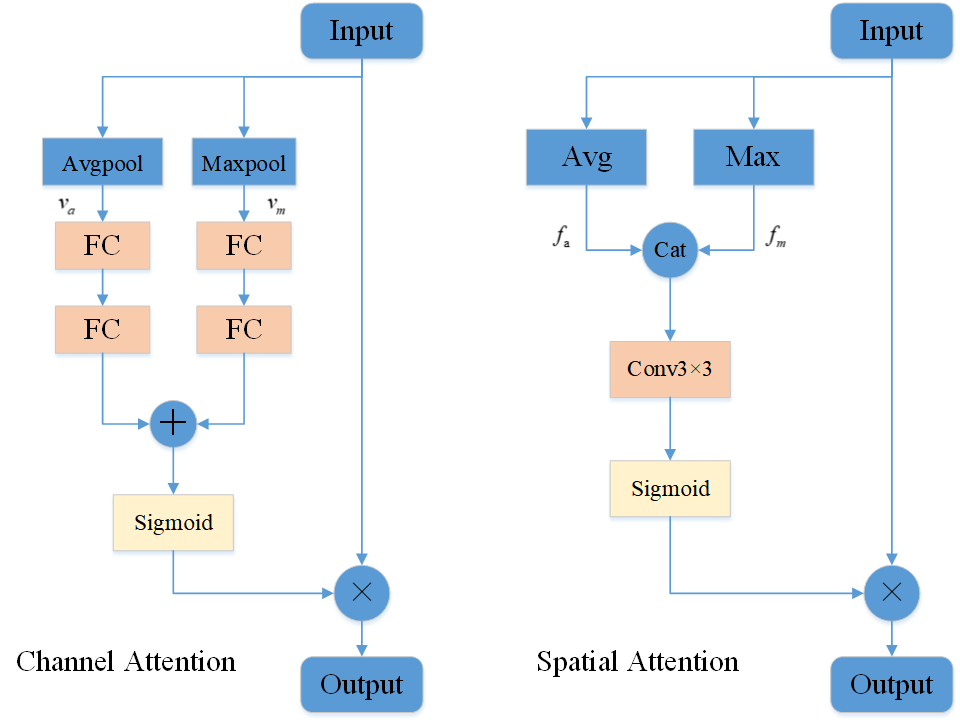}
  \caption{The architecture of channel attention module and spatial attention module.} 
\end{figure}

\subsection{Attention-based pyramid feature aggregation (APFA) mechanism}
In many networks based on the U-Net model, the final predicted saliency map comes from the last layer of the top-down decoder. However, the truth is that the deep layer features will be gradually diluted when they are transmitted into shallow layers \cite{liu2019simple}. Hence, we  generate the final saliency map from these four sets of multi-scale contextual feature maps together. But another problem arises that there is a semantic gap between different levels of multi-scale contextual feature maps, so simply combining them will lead to sub-optimal results. Based on the above discussion, we introduce an attention-based pyramid feature aggregation (APFA) mechanism to efficiently fuse multi-scale information of multi-layer features, and use the channel attention module and spatial attention module for deep and shallow pyramid features, respectively.

The proposed APFA module is shaped like an inverted pyramid as shown in Fig. 1. Inspired by WFNet \cite{cen2020wfnet} and MINet \cite{pang2020multi}, the feature maps of two adjacent layers are combined two by two to generate the pyramid feature maps of next layer, thus effectively alleviating the problem of semantic gap. In addition, we find that the feature map with the smallest resolution in each layer of the pyramid is the deep pyramid feature, while the feature map with the largest resolution is the shallow pyramid feature. Therefore, for the deep features, the channel attention module is adopted to model the semantic information. And for the shallow features, the spatial attention module is designed to weight the spatial information. The architectures of channel and spatial attention module are shown in Fig. 3.

\textbf{Channel attention module.} We use the channel attention module for the deep pyramid feature map. Specifically, the input feature map $F$ is transformed into vectors $v_a$ and $v_m$ after the average pooling and maximum pooling in the spatial dimension, respectively. We then feed these two vectors into a two-layer fully connected layer. Finally, the two feature vectors are combined as a channel attention map and its values are constrained to be between $[0,1]$. In Fig. 3, this process can be formulated as:
\begin{equation}
F_c = \sigma (fc(fc(avgpool(F))) + fc(fc(maxpool(F)))) \odot F
\end{equation}

\textbf{Spatial attention module.} Since shallow feature maps with larger resolution contain more detail information, not all of these features  belong to the salient objects. Hence we use the spatial attention module to highlight salient regions and suppress background interference. Firstly, the input feature map $F$ is transformed into two single-channel feature maps $f_a$ and $f_m$ by average pooling and maximum pooling in the channel dimension, respectively. Then $f_a$ and $f_m$ are concatenated and fused to obtain the final spatial attention map with values between $[0,1]$. In Fig. 3, this process can be formulated as:
\begin{equation}
  F_s = \sigma (Conv_{3 \times 3} (Cat(avg (F),max(F)))) \odot F
\end{equation}
in which $avg$ and $max$ are the average-pooling and max-pooling in channel dimension.

\subsection{Loss function}
In SOD task, the binary cross-entropy loss is the most commonly used for network training. And the standard binary cross-entropy can be formulated as follows:
\begin{equation}
  Loss_{bce}  = \sum\limits_{p \in P,g \in G} { - [g\log p + (1 - g)\log (1 - p)]} 
\end{equation}
where $P$ is the predicted saliency map, $G$ is the saliency ground truth label, $p$ denotes the probability of belonging to a salient object and the value of $g$ is equal to 0 or 1.

As mentioned in BASNet \cite{qin2019basnet}, IoU is a map-level measure for object detection and segmentation. And it has been used as the training loss for SOD task. Hence, we introduce the IoU loss as the training loss. The IoU loss function can be formulated as follows:
\begin{equation}
Loss_{iou}  = 1 - \frac{{(\sum\limits_{p \in P,g \in G} {pg} ) + 1}}{{(\sum\limits_{p \in P,g \in G} {(p + g) - pg} ) + 1}}
\end{equation}

In the end, the whole loss function can be written as:
\begin{equation}
Loss  = Loss_{bce}  + \lambda Loss_{iou} 
\end{equation}
where $\lambda$ is a hyperparameter that balances the contributions of the two losses. Empirically, we set $\lambda  = 0.6$.

\section{Experiment}
\subsection{Datasets and Evaluation Metrics}
The existing datasets for remote sensing image salient object detection are ORSSD and EORSSD. The ORSSD dataset contains 800 challenging remote sensing images and their corresponding pixel-level annotated truth maps, 600 images as the training set and 200 images as the test set. The EORSSD dataset is an expansion of the ORSSD dataset by adding more semantically meaningful and challenging images, with a total of 2000 images, 1400 as the training set and 600 as the test set. We follow the previous work \cite{zhang2020dense} and use the training set of EORSSD to train our network and evaluate the model on the test set of ORSSD and EORSSD.

For quantitative evaluation, four widely used metrics including mean absolute error ($MAE$), maximum F-measure ($maxF$), 
structure-measure ($Sm$) \cite{fan2017structure} and enhanced-alignment measure ($Em$) \cite{fan2018enhanced} are used to evaluate the performance of MSCNet and existing state-of-the-art methods.

\subsection{Implementation Details}
The model is totally trained for 40 epochs on the training set of EORSSD with 1400 images. Random flipping and  random rotating act as 
data augmentation techniques to improve the diversity of training set. MobileNet V2, pre-trained on ImageNet, is used as the backbone encoder network. We train the proposed MSCNet using the Adam optimizer with initial learning rate of 1e-4, weight decay of 5e-4, and batch size of 6. Cosine annealing decay strategy is adopted to adjust the learning rate. PyTorch 1.8.1 is employed to implement our model and a GeForce GTX Titan X GPU is used for acceleration. And the input size is $224 \times 224$.

\subsection{Ablation Studies}
In this section, we perform a series of ablation experiments to investigate the importance of different modules proposed in this paper. Our experiments are based on the same and fair experimental setup. The qualitative comparison and quantitative comparison are shown in Fig. 4 and Table. \uppercase\expandafter{\romannumeral1}. 
The baseline is a U-shaped network that uses MobileNet V2 as the backbone and uses four simple convolutional layers to perform multi-layer feature fusion.

\textbf{Analysis of MSCE.} Compared with the baseline, due to the introduction of the MSCE module as shown in Table. \uppercase\expandafter{\romannumeral1}, the F-measure is boosted from 0.8596 to 0.8729 with the percentage gain of 1.33\%. And the E-measure is improved from 0.9524 to 0.9609 with a percentage gain of 0.85\%. In addition, it can be seen that the proposed module works better on detecting and locating multi-scale salient objects. For example, the two examples shown in rows 1 and 2 in Fig. 4 are challenging scenarios in EORSSD, i.e., a large number of small objects with a small occupancy scale. Before equipped with the MSCE module, the model could neither describe the correct position of the buildings and vehicles nor accurately locate the boundaries of the small objects. However, as can be seen from Fig. 4 (d), the MSCE module significantly improves the network's ability to detect multi-scale objects, such as more precise location of the salient objects in row 2 and buildings that fit more closely to the boundary in row 1.

\textbf{Analysis of APFA.} In Table. \uppercase\expandafter{\romannumeral1}, we also demonstrate the effectiveness of the APFA module. In terms of F-measure, the APFA module brings a gain of 0.74\% to the baseline. And the APFA module contributes 0.68\% of a percentage point to the baseline in terms of E-measure. As shown in Fig. 4(e), thanks to the APFA module's progressive fusion of multi-layer contextual feature maps and the use of the attention module, the network can eventually detect objects with large scale ranges like rivers in row 3, while the baseline can barely detect any relevant features of the river.
\renewcommand\arraystretch{1.5}
\begin{table}[!ht]
  \centering
  \caption{Quantitative evaluation of ablation studies on the test set of EORSSD dataset.}
  \begin{tabular}{c|c|c}
  \hline
      \multirow{2}{*}{Methods} & \multicolumn{2}{c}{EORSSD} \\ \cline{2-3}
      ~ & $maxF\uparrow$ & $E_m\uparrow$ \\ \hline
      Baseline & 0.8596 & 0.9524 \\ \hline
      Baseline+MSCE & 0.8729 & 0.9609 \\ \hline
      Baseline+APFA & 0.8670 & 0.9592 \\ \hline
      Baseline+MSCE+APFA & 0.8806 & 0.9684 \\ \hline
  \end{tabular}
\end{table}
\begin{figure}[h]
  \centering
  \includegraphics[width=3in]{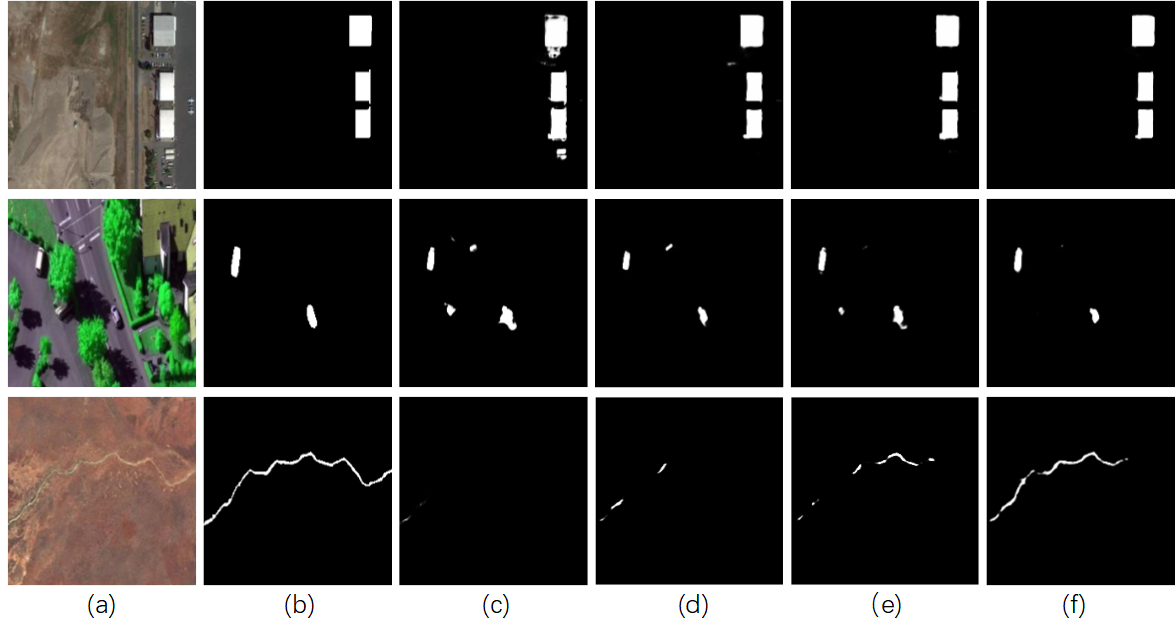}
  \caption{Comparison of predictions among different networks. (a) Image; (b) Ground truth; (c) Baseline; (d) Baseline+MSCE; (e) Baseline+APFA; (f) Baseline+MSCE+APFA.} 
\end{figure}

In summary, both of the proposed key modules further enhance the performance of the network. Specifically, MSCNet achieves a gain of 2.1\% in F-measure and a gain of 1.6\% in E-measure compared with the baseline. And the combination of these two modules provides the best visual results as shown in Fig. 4(f).

\subsection{Comparison with State-of-the-arts}
To demonstrate the validity of our proposed network, we compare the proposed method against other fifteen current state-of-the-art (SOTA) methods, including eleven SOD methods for NSIs (i.e., R3Net\cite{deng2018r3net}, DSS\cite{hou2017deeply}, RADF\cite{hu2018recurrently}, PFAN\cite{zhao2019pyramid}, PoolNet\cite{liu2019simple}, EGNet\cite{zhao2019egnet}, F3Net\cite{wei2020f3net}, ITSD\cite{zhou2020interactive}, LDF\cite{wei2020label}, MINet\cite{pang2020multi} and GCPANet\cite{chen2020global}), two lightweight SOD methods for NSIs (i.e., SAMNet\cite{liu2021samnet} and HVPNet\cite{liu2020lightweight}), and two recent SOD methods for optical RSIs (i.e., LVNet\cite{li2019nested}, and DAFNet\cite{zhang2020dense}). For a fair comparison, all of the compared saliency maps are provided by \cite{zhang2020dense} or generated by the source codes provided by the authors. All SOD methods designed for NSIs are retrained on the same dataset as the method in this paper and use the original setup parameters of the corresponding method. Table. \uppercase\expandafter{\romannumeral2} and Fig. 5 show the comparison of the different methods on the four metrics and visualization results, respectively.

\begin{figure*}[ht]
  \centering
  \includegraphics[width=5in]{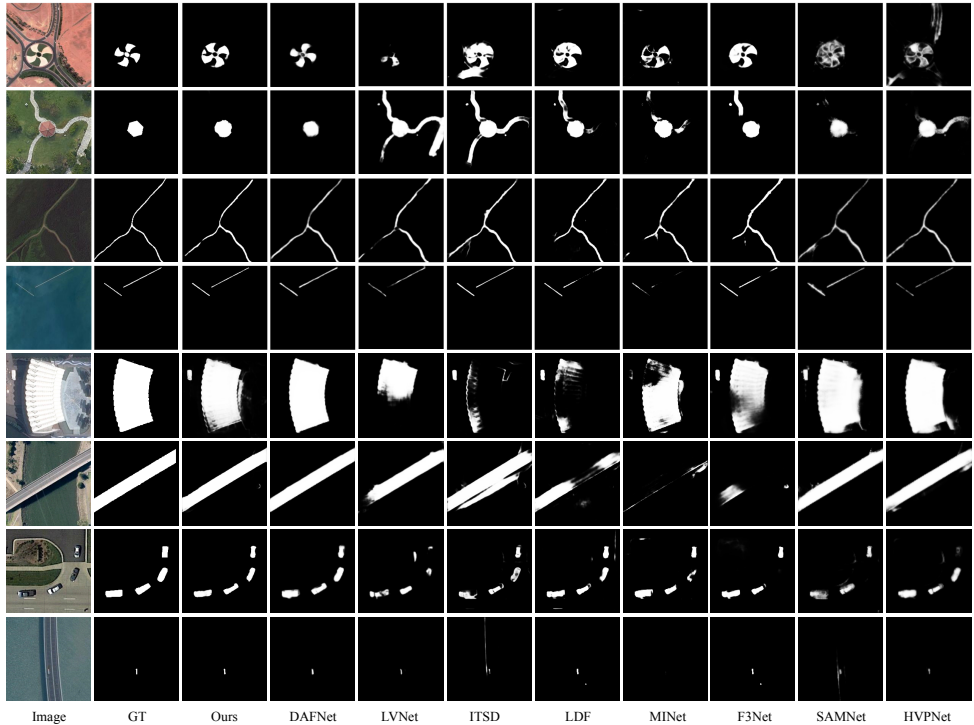}
  \caption{Visual comparisons of different methods.} 
  \label{SOTA}
\end{figure*}
\renewcommand\arraystretch{1.5}
\begin{table}[ht]
  \centering
  \caption{Quantitative evaluation. The best three results are highlighted in red, green and blue.}
  \setlength{\tabcolsep}{1mm}{
  \begin{tabular}{|c|c|c|c|c|c|c|c|c|}
  \hline
      \multirow{2}{*}{Methods}  & \multicolumn{4}{c|}{ORSSD} & \multicolumn{4}{c|}{EORSSD} \\ \cline{2-9}
      ~  & $maxF$ & MAE & $S_m$ & $E_m$ & $maxF$ & MAE & $S_m$ & $E_m$ \\ \hline
      R3Net  & .7612 & .0403 & .8171 & .8854 & .7751 & .0174 & .8192 & {\color{blue}.9496} \\ \hline
      DSS  & .7621 & .0366 & .8261 & .8857 & .7067 & .0188 & .7868 & .9196 \\ \hline
      RADF  & .7776 & .0384 & .8256 & .9130 & .7743 & .0167 & .8211 & .9182 \\ \hline
      PFAN  & .8222 & .0250 & .8542 & .9481 & .7609 & .0158 & .8362 & .9157 \\ \hline
      PoolNet  & .7841 & .0360 & .8396 & .9311 & .7560 & .0208 & .8214 & .9165 \\ \hline
      EGNet  & .8476 & .0219 & .8725 & {\color{blue}.9722} & .7988 & {\color{blue}.0113} & .8593 & .9398 \\ \hline
      F3Net  & .8758 & .0199 & .8985 & .9498 & {\color{blue}.8543} & .0118 & {\color{blue}.8887} & .9485 \\ \hline
      ITSD  & {\color{blue}.8795} & {\color{blue}.0184} & {\color{blue}.8987} & .9619 & .8392 & .0117 & .8803 & .9448 \\ \hline
      LDF  & .8627 & .0229 & .8859 & .9400 & .8527 & .0140 & .8799 & .9448 \\ \hline
      MINet  & .8585 & .0205 & .8856 & .9353 & .8206 & .0119 & .8753 & .9215 \\ \hline
      GCPANet  & .8635 & .0214 & .8801 & .9458 & .8000 & .0132 & .8590 & .9178 \\ \hline
      SAMNet  & .8516 & .0218 & .8828 & .9512 & .7791 & .0151 & .8518 & .9192 \\ \hline
      HVPNet  & .8422 & .0202 & .8848 & .9455 & .7855 & .0127 & .8600 & .9247 \\ \hline
      LVNet  & .8434 & .0211 & .8807 & .9454 & .8071 & .0146 & .8639 & .9274 \\ \hline
      DAFNet  & {\color{red}.9192} & {\color{red}.0111} & {\color{green}.9186} & {\color{red}.9822} & {\color{red}.8985} & {\color{red}.0060} & {\color{red}.9175} & {\color{red}.9805} \\ \hline
      MSCNet  & {\color{green} .9107} & {\color{green} .0132} & {\color{red}.9226} & {\color{green} .9754} & {\color{green}.8806} & {\color{green}.0090} & {\color{green}.9086} & {\color{green}.9684} \\ \hline
  \end{tabular}}
\end{table}

\renewcommand\arraystretch{1.5}
\begin{table}[ht]
  \centering
  \caption{Comparison of the parameters. Rows 1 and 3 are the specific methods, rows 2 and 4 are the corresponding parameters.}
  \setlength{\tabcolsep}{1mm}{
  \begin{tabular}{c|c|c|c|c|c|c}
  \hline
      R3Net & DSS  & PFAN & PoolNet & EGNet & F3Net & ITSD \\ \hline
      56.16M & 62.23M  & 16.38M & 53.63M & 108.07M & 25.54M & 26.47M \\ \hline
      LDF & MINet & GCPANet & SAMNet & HVPNet  & DAFNet & MSCNet \\ \hline
      25.15M & 47.56M & 67.06M & 1.33M & 1.23M  & 29.35M & 3.26M \\ \hline
  \end{tabular}}
\end{table}

\textbf{Quantitative comparison}. Table. \uppercase\expandafter{\romannumeral2} shows the evaluation scores of the method in this paper and other SOTA methods on F-measure, MAE score, S-measure, and E-measure. Except for MAE score, which is the smaller the better, all other metrics are the larger the better.
Moreover, the parameters of these models are also displayed in Table. \uppercase\expandafter{\romannumeral3}. As shown in Table. \uppercase\expandafter{\romannumeral2}, the proposed MSCNet beats all SOTA methods designed for NSI SOD task which demonstrates the necessity to design network models specially for optical RSIs. For example, our method achieves a percentage gain of 2.63\% and 3.12\% in terms of F-measure compared to the best NSI SOD method on EORSSD and ORSSD, respectively. In addition, it is worth noting that the model size of our method is much smaller than that of the best method DAFNet. By analyzing the comparison on all the metrics, our method achieves 98\% performance on F-measure, 99\% performance on S-measure, 98.8\% performance on E-measure on the EORSSD dataset with only 11.1\% of the parameters of DAFNet. The above results clearly show that our lightweight model achieves comparable accuracy with state-of-the-art SOD solutions.

\textbf{Qualitative comparison}. In Fig. 5, we provide the visual comparison to exhibit the superiority of the proposed MSCNet. The first column is the images and the second column is the corresponding ground truths. Our result is in the third column. As can be seen from rows 1 and 2, our method can accurately identify the location of salient objects and suppress redundant noise, which shows that MSCNet performs much better than other SOTA methods even when there is a complex background around the salient objects. And for the slender difficult samples, MSCNet also achieves the optimal detection results with more complete saliency map in the rows 3 and 4 in Fig. 5, as the APFA module carefully refines and fuses multi-layer contextual information. In addition, the proposed method shows outstanding robustness in dealing with multi-scale objects. For example, the small car in the last image and the buildings in row 5. And we think that this should be the contribution of MSCE. The above results fully confirm the effectiveness of the proposed method for RSI SOD task.

\section{Conclusion}
In this paper, a novel end-to-end lightweight salient object detection network built specially for optical RSIs is proposed, named MSCNet. Benefiting from the proposed multi-scale context extraction module, our network can sufficiently extract multi-scale contextual information to accurately locate and segment salient objects at different scales. Moreover, The proposed attention-based pyramid feature aggregation mechanism employs a hierarchical fusion approach to further refine and complement the saliency map when gradually fusing the multi-layer contextual features. Due to the use of the attention mechanism, this module also further enhances the modeling of semantic information and the suppression of background interference to produce optimal results. In the future, we intend to improve the effectiveness of our lightweight model for accurate segmentation of salient object boundaries with the help of edge detection related methods.

\section*{Acknowledgment}
This work is supported in part by the Fundamental Research Funds for the Central Universities of China under Grant NZ2019009.

\bibliographystyle{IEEEtran}
\bibliography{IEEEabrv,bare_conf.bib}

\end{document}